\documentclass[letterpaper, 10 pt, conference]{ieeeconf}  % Comment this line out if you need a4paper

\IEEEoverridecommandlockouts                              % This command is only needed if 
\usepackage{graphicx}
\usepackage{amsmath}
\usepackage{amssymb}
\usepackage{subcaption}
\usepackage[normalem]{ulem}
\usepackage{color}
\usepackage{xcolor}

\title{\LARGE \bf
Symbolic Manipulation Planning with Discovered Object and Relational Predicates
}

\author{Alper Ahmetoglu$^{1}$, Erhan Oztop$^{2,3}$, Emre Ugur$^{1}$% <-this % stops a space
\thanks{$^{1}$Department of Computer Engineering, Bogazici University
        {\tt\small alper.ahmetoglu@boun.edu.tr}}%
\thanks{$^{2}$Department of Computer Science, Ozyegin University}%
\thanks{$^{3}$OTRI, SISReC, Osaka University}
}

\begin{document}

\maketitle
\pagestyle{empty}

%%%%%%%%%%%%%%%%%%%%%%%%%%%%%%%%%%%%%%%%%%%%%%%%%%%%%%%%%%%%%%%%%%%%%%%%%%%%%%%%
\begin{abstract}
Discovering the symbols and rules that can be used in long-horizon planning from a robot's unsupervised exploration of its environment and continuous sensorimotor experience is a challenging task. The previous studies proposed learning symbols from single or paired object interactions and planning with these symbols. In this work, we propose a system that learns rules with discovered object and relational symbols that encode an arbitrary number of objects and the relations between them, converts those rules to Planning Domain Description Language (PDDL), and generates plans that involve affordances of the arbitrary number of objects to achieve tasks. We validated our system with box-shaped objects in different sizes and showed that the system can develop a symbolic knowledge of pick-up, carry, and place operations, taking into account object compounds in different configurations, such as boxes would be carried together with a larger box that they are placed on. We also compared our method with the state-of-the-art methods and showed that planning with the operators defined over relational symbols gives better planning performance compared to the baselines.
\end{abstract}

%%%%%%%%%%%%%%%%%%%%%%%%%%%%%%%%%%%%%%%%%%%%%%%%%%%%%%%%%%%%%%%%%%%%%%%%%%%%%%%%
\section{INTRODUCTION}
\label{sec:intro}
\begin{figure*}[htbp]
    \centering
    \includegraphics[width=0.98\textwidth]{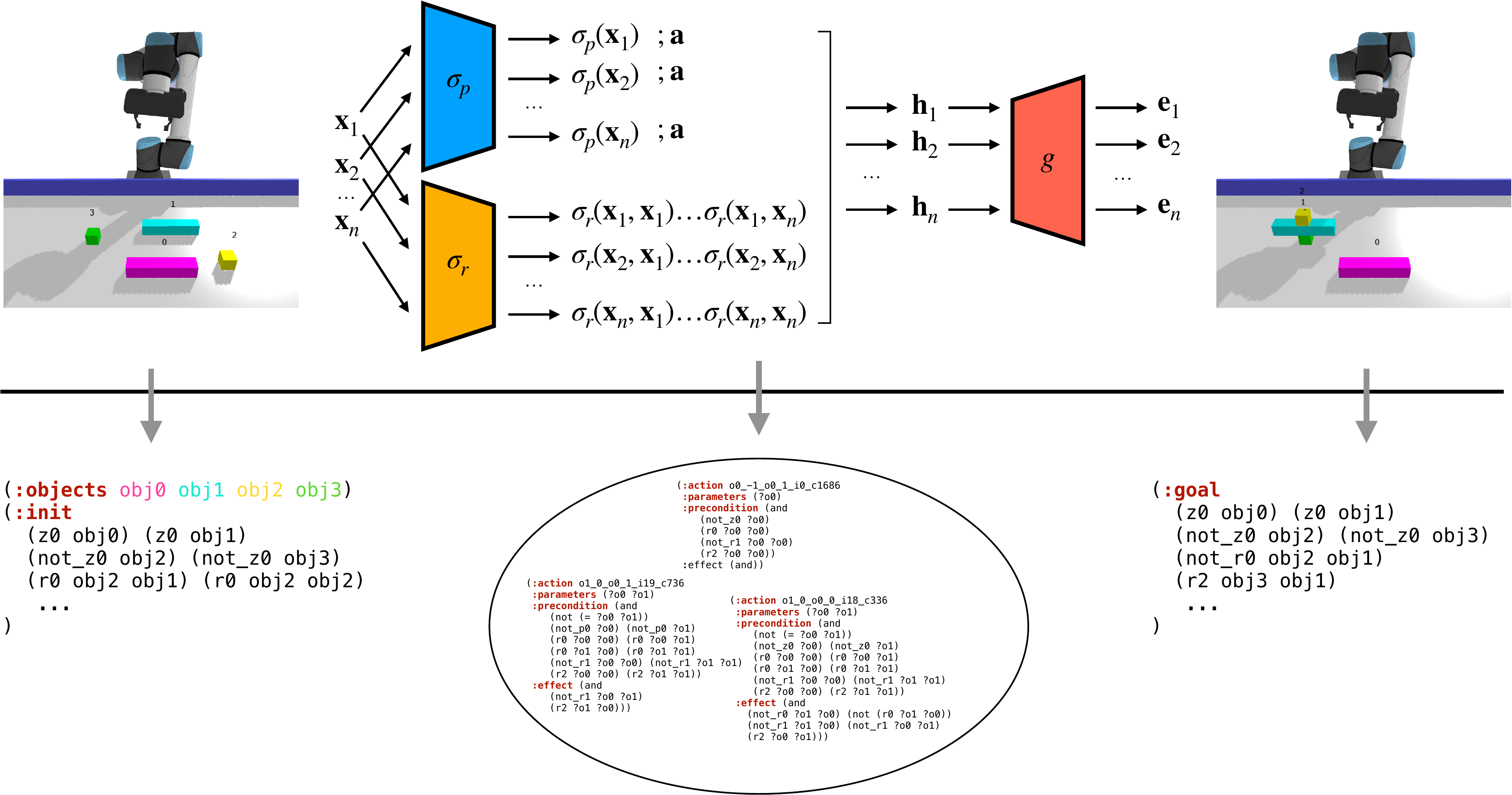}
    \caption{An overview of the proposed method. (Top) The encoder network $\sigma_p$ learns unary symbols over object features while the self-attention network $\sigma_r$ learns relational symbols. The aggregation in the middle (see Equation \ref{eq:agg}) fuses unary and relational symbols with action into a vector representation for each object, which is given as input to the decoder network $g$ to predict the effect of the executed action. (Bottom) After learning operators defined over the learned symbols, we can find a sequence of actions that reach the goal state from the initial state with AI planners.}
    \label{fig:model}
\end{figure*}
A long-standing challenge for artificial intelligence (AI) research is to build a generalist agent that can parse and understand its environment through its sensors, carry out desired tasks by acting on the environment, and update its knowledge when necessary to adapt to new situations. The difficulty in achieving this stems from the fact that we do not have a well-defined, robust, and generic representation of the environment that applies to a sufficiently large class of tasks. The requirements and the resolution of the representation change drastically when the aimed task properties and/or the capabilities of the agent change. For example, a robotic agent tidying up the table needs a different representation of the environment than a robot tying a rope. It is not even clear whether soft and rigid objects should be represented within a single representational framework. Nevertheless, once a symbolic representation of the environment that is appropriate for the task at hand can be obtained, many AI search techniques become available for efficiently finding a solution to problems such as planning to achieve a goal \cite{konidaris2019necessity}.

An effective approach to build a set of symbols for obtaining the leverage of off-the-shelf symbolic systems is to focus on the preconditions and effects related to the actions of an agent \cite{konidaris2014constructing,konidaris2018skills,Ugur-2015-ICRA}. Learning precondition-effect relations effectively models the environment based on the agent's capabilities. This is preferable for filtering out irrelevant information that would otherwise increase the complexity of the problem. Once the necessary symbols are learned, the environment description can be translated, for example, into Planning Domain Definition Language (PDDL) \cite{aeronautiques1998pddl} that allows the use of fast domain-independent planners such fast downward \cite{helmert2006fast} or fast-forward \cite{hoffmann2001ff}. Alternatively, one can first compress the experience of the agent, i.e., the sensory state-space explored, into a symbolic space using deep neural networks, then encode symbolic transitions either with PDDL or learn a separate module for generating the necessary action from the current state for a given goal state \cite{asai2018classical,asai2020learning,asai2022classical}. One advantage of this approach is that it can be integrated with other types of deep architectures to process different of sensory information, such as images, sounds, or proprioceptive information.

Our previous work, DeepSym \cite{ahmetoglu2022deepsym}, stands in between these two approaches by having a differentiable, deep architecture that can learn symbolic representations for the preconditions of the executed action. DeepSym is composed of an encoder-decoder network with binary bottleneck layers for learning object symbols to predict the effect of the executed action. In a follow-up work \cite{ahmetoglu2022learning}, the architecture is improved by employing a transformer layer in the decoder, allowing symbols to interact to model multi-object effects. However, the interaction of symbols remains implicit in the weights of the transformer and cannot be translated into PDDL in a straightforward way, thus limiting the domain-independent planning capability. In other words, even though the architecture models relations between objects, these relations cannot be expressed as relational symbols between objects. Our recent work \cite{ahmetoglu2023discovering} proposes a solution for this problem with an architecture that can learn not only unary object symbols but explicitly encodes relations between objects using binary attention weights. However, symbolic-level transitions from the learned symbols that enable domain-independent planning were not considered.

In the current study, we first learn a set of unary and relational symbols between objects using the Relational DeepSym architecture \cite{ahmetoglu2023discovering}, then we propose a method to build abstract operators defined over these unary and relational symbols that describe the symbolic transition of a state when an action is executed. We translate these operators into PDDL descriptions and show that they can be used for planning with off-the-shelf AI planners in a tabletop object stacking task. We compare our method with \cite{ahmetoglu2022deepsym} and \cite{ahmetoglu2022learning} in terms of effect prediction accuracy and planning performance. Our results show that planning with the operators defined over relational symbols performs better than the baselines.

\section{RELATED WORK}
\label{sec:related}
There is a large body of literature where the learned affordances or effect predictors have been used to make plans in the continuous or sub-symbolic space of the robots \cite{ugur2011unsupervised,Zech2017}. Our work is mainly related to methods that study symbol emergence \cite{taniguchi2016symbol} and that discover symbolic representations of the environment from the continuous experience of the robot through its sensors with the aim of planning with off-the-shelf AI planners.

\cite{konidaris2014constructing,konidaris2018skills} showed that it is necessary and sufficient to learn symbols for the precondition and the effect set of the agent's action repertoire to enable planning. Based on this observation, they learn a set of symbols that cover actions' pre- and post-conditions and use them to build a PDDL description of the environment. These works use a global state representation in which each object is represented in the global state vector, restricting the use of previously learned symbols. In follow-up work, \cite{james2020learning} builds upon this framework by using an ego-centric state representation, allowing agent-centric symbols to be shared through different tasks. \cite{Ugur-2015-ICRA,Ugur-2015-Humanoids} considers object-centric precondition and effect set to find discrete symbols for producing PDDL descriptions. Similarly, \cite{james2021autonomous} used object-centric representations, which increases the generalization of symbols through objects with the same properties. Our work differs from these in that we learn symbols with deep neural networks by minimizing the effect prediction error instead of clustering the state space.

In another line work, \cite{silver2021learning,chitnis2022learning,silver2022learning} proposed a bi-level planning schema in which a set of operators and corresponding samplers are learned from previously acquired symbols that allow the agent to make refined plans that can consider the geometric information. \cite{silver2023predicate} learns predicates from demonstrations with a surrogate objective for planning. \cite{achterhold2023learning} learns policies parameterized by neural networks based on the symbolic state transitions. Quite similar to our motivation, \cite{kumar2023learning} learn a set of operators by considering only a subset of abstract effects to prevent learning complex operators. The main difference between our work and theirs is that we select this subset as the relevant symbols for the action, whereas they consider universal quantifiers in the abstract effect.

\cite{asai2018classical,asai2020learning,asai2022classical} use a state autoencoder to compress the state space into a low-dimensional binary vector and an action autoencoder to learn action representations from low-level state transitions. Similarly, \cite{ahmetoglu2022deepsym} uses an encoder-decoder style architecture to generate symbolic representations in the bottleneck layer while minimizing the effect prediction error. In follow-up work, \cite{ahmetoglu2022learning} employed transformer layers \cite{vaswani2017attention}, enabling the flow of information between object symbols to model multi-object effects. Recently, \cite{ahmetoglu2023discovering} proposed a relational formulation of the encoder-decoder architecture that explicitly models relations between objects using binary attention weights, learning unary and relational symbols simultaneously. Our work builds upon this architecture by proposing a method to learn operators defined over these symbols that can be used for planning with AI planners.

\section{METHOD}
\label{sec:method}

\subsection{Problem Definition}
\label{subsec:problem}
This paper deals with the problem of learning (1) a symbolic representation of an environment defined over a finite set of objects, which is sensed as a  continuous sensory state, and (2) a set of operators that represent the dynamics of sensory transition dynamics at the symbolic level, which allows domain-independent planning with the learned symbols by the use of off-the-shelf AI planners.

An environment is characterized by a tuple $(\mathcal{X}, \mathcal{A}, P)$ where $\mathcal{X}$ denotes the continuous sensory state of the environment (which is called state-space from now on), $\mathcal{A}$ is a finite set of actions that the agent can execute, and $P(\mathbf{X}'\mid \mathbf{X}, \mathbf{a})$ is the probability that taking action $\mathbf{a} \in \mathcal{A}$ at state $\mathbf{X}$ results in state $\mathbf{X}'$. An environment instance consists of a set of objects $\{o_1, \dots, o_n\} \in \mathcal{O}$ each having a $d_o$-dimensional continuous-valued feature vector $\mathbf{x}_i \in \mathbb{R}^{d_o}$ defining the state of the environment as an unordered set of feature vectors $\mathbf{X} := \{\mathbf{x}_1, \dots, \mathbf{x}_n\} \in \mathcal{X}$. $\mathbf{X}$ is not a fixed-size vector but a variable size depending on the number of objects $n$ in the environment. An action $\mathbf{a}$ is a fixed-size vector representing a high-level parameterized movement primitive that the agent can execute, such as picking an object. Given an initial state $\mathbf{X}_0$ and a goal state $\mathbf{X}_g$, the objective is to find a sequence of actions $(\mathbf{a}_1, \dots, \mathbf{a}_k)$ that maximizes the probability of reaching the goal state.

We are interested in learning the mapping $f : \mathcal{X} \rightarrow \mathcal{P}$ that transforms a state vector $\mathbf{X}$ into a set of predicates (or symbols) $\Sigma := \{\sigma_1(\mathbf{X}), \dots, \sigma_m(\mathbf{X})\} \in \mathcal{P}$, where $\sigma_i : \mathcal{X} \rightarrow \{0, 1\}^{d_k}$ is a binary function where $d_k$ is an environment dependent fixed dimension. After symbols are learned, we can find a set of operators (i.e., lifted actions in the symbolic space) $\{\phi_1, \dots, \phi_k\} \in \Phi$ in which each operator $\phi_i : \Sigma \rightarrow \Sigma$ transforms the current symbols into a new set of symbols. Once symbols and operators are learned, we can transform the initial state $\mathbf{X}_0$ and the goal state $\mathbf{X}_g$ into symbolic representations $\Sigma_0$ and $\Sigma_g$, respectively, and then find a sequence of operators $(\phi_{1}, \dots, \phi_{k})$ that transforms $\Sigma_0$ into $\Sigma_g$, and then execute the corresponding sequence of actions $(\mathbf{a}_{1}, \dots, \mathbf{a}_{k})$ to reach the goal state.

Note that the proposed operator learning system is built on top of our symbol learning network architecture \cite{ahmetoglu2023discovering}. Thus, to assess the added value of learning explicit symbolic transitions,  we compare the performance of the proposed model in effect prediction and planning with \cite{ahmetoglu2022deepsym} and \cite{ahmetoglu2022learning} in a tabletop object manipulation setup.

\subsection{Learning Unary and Relational Symbols}
\label{subsec:learning}
Figure \ref{fig:model} shows an outline of the method. The architecture is composed of four main blocks.

\subsubsection{Encoder Network}
\label{subsubsec:encoder}
$\sigma_p: \mathcal{X} \rightarrow \mathcal{P}$ is a multi-layer perceptron (MLP) with Gumbel-Sigmoid (GS) \cite{maddison2016concrete} activation that outputs a binary number for each object in the environment. We treat this as a unary predicate $\sigma_{\text{p}}(\mathbf{x}_i)$ that encodes the property of an object. The number of properties that can be encoded is bounded by the output dimensionality of the MLP -- at most $2^{\text{d}}$ properties can be encoded with $d$-dimensional outputs.

\subsubsection{Self-Attention Network}
\label{subsubsec:attention}
$\sigma_r: \mathcal{X} \rightarrow \mathcal{P}$ is a multi-layer perceptron combined with a modified version \cite{ahmetoglu2023discovering} of the original self-attention layer \cite{vaswani2017attention}. Given a state vector $\mathbf{X}=\{\mathbf{x}_1, \dots, \mathbf{x}_n\}$, the MLP part outputs a set of $d$-dimensional vectors $\mathbf{Q}=\{\mathbf{q}_1, \dots, \mathbf{q}_n\}$ and $\mathbf{K}=\{\mathbf{k}_1, \dots, \mathbf{k}_n\}$ where $\mathbf{q}_i$ and $\mathbf{k}_i$ are the query and key vectors for object $o_i$, respectively. Unlike the original self-attention layer, we do not define value vectors as we are interested in the attention values. The second important difference is the computation of the attention values:
\begin{equation}
    \alpha_{ij} = \text{GumbelSigmoid}(\mathbf{q}_i \cdot \mathbf{k}_j)
\end{equation}
Firstly, using the sigmoid function instead of softmax allows attention values to focus on multiple tokens independently. Secondly, the binarization (due to GS) of attention values allows us to treat them as binary relations between objects while preserving differentiability.

We define the output of the whole block as a relational predicate $\sigma_{\text{r}}(\mathbf{x}_i, \mathbf{x}_j)$ that encodes the relation between objects $o_i$ and $o_j$. Note that the order of arguments is important since  $\mathbf{q}_i \cdot \mathbf{k}_j$ and $\mathbf{q}_j \cdot \mathbf{k}_i$ can have different values. Note that, without loss of generality, we described the operation with a single attention head; however, in the implementation, we used three attention heads.

\subsubsection{Aggregation Function}
\label{subsubsec:aggregation}
fuses unary predicates $\sigma_{\text{p}}(\mathbf{x}_i)$, relational predicates $\sigma_{\text{r}}(\mathbf{x}_{i}, \mathbf{x}_j)$, and the action vector $\mathbf{a}$ in a single representation $\mathbf{h}_i$ for each object that is fed into the decoder network for predicting the effect of the executed action. The aggregation function is defined as follows:
\begin{align}
    \mathbf{z}_i &= \text{MLP}([\sigma_{\text{p}}(\mathbf{x}_i); \mathbf{a}]) & \forall i \in \{1, \dots, n\} \\
    \mathbf{h}_i^k &= \sum_{j=1}^{n} \sigma_{\text{r}_k}(\mathbf{x}_i, \mathbf{x}_j) \mathbf{z}_i & \forall i \in \{1, \dots, n\}\label{eq:agg}\\
    && \forall k \in \{1, \dots, K\}\\
    \mathbf{h}_i &= [\mathbf{h}_i^1; \dots; \mathbf{h}_i^K]
\end{align}
where $K$ is the number of relation types. The action vector $\mathbf{a}$ is concatenated with the unary predicate $\sigma_{\text{p}}(\mathbf{x}_i)$, and then fed into an MLP to obtain a representation $\mathbf{z}_i$ that holds action information. Then, for each relation $k$ and object $i$, intermediate representations $\mathbf{z}_j$ are summed up for indices that satisfy the relation $\sigma_{\text{r}_k}(\mathbf{x}_i, \mathbf{x}_j)$, resulting in a fixed-size vector $\mathbf{h}_i^k$ containing information regarding objects that have a relation $r_k$ with object $o_i$. Lastly, $\{h_i^1, \dots, h_i^k\}$ are concatenated to obtain the aggregated representation $\mathbf{h}_i$ that can hold any necessary information about the object $o_i$, the action $\mathbf{a}$, and the relations between $o_i$ and other objects. Equation \ref{eq:agg} essentially enables message passing between different object symbols based on the learned relations between objects.

\subsubsection{Decoder Network}
\label{subsubsec:decoder}
$g$ is an MLP that takes the aggregated representation $\mathbf{h}_i$ as input and outputs the effect $\hat{\mathbf{e}}_i$ of action $\mathbf{a}$ on the object $o_i$. The predicted effect is used to compute the mean squared error that is backpropagated through the whole network:
\begin{equation}
    \mathcal{L} = \sum_{i=1}^{n} \lVert \hat{\mathbf{e}}_i - \mathbf{e}_i \rVert^2
    \label{eq:loss}
\end{equation}
where $n$ is the number of objects and the effect vector $\mathbf{e}_i$ is defined as the difference between the current state $\mathbf{x}_i$ and the next state $\mathbf{x}_i'$.

These four blocks create a single differentiable module that can learn unary and relational predicate symbols over object features to minimize the effect prediction error in an end-to-end fashion. To train the network, we execute random actions in the environment and collect a dataset of $(\mathbf{X}, \mathbf{a}, \mathbf{X}')$ tuples. Then, we train the network to minimize the effect prediction error $\mathcal{L}$ defined in Equation \ref{eq:loss}.

\subsection{Learning Operators}
\label{subsec:ops}
In this section, we describe how to learn operators that can be used for planning. Throughout this section, we will denote a ground symbol as $\sigma_p{(\mathbf{x}_i)}$, a lifted symbol as $\sigma_p(?x)$, and a substitution as $\theta=\{?x/\mathbf{x}_i\}$ where $?x$ indicates a free variable that is not bound to any object.

After we train the network, we can transform the dataset $\{(\mathbf{X}^{(i)}, \mathbf{a}^{(i)}, \mathbf{X'}^{(i)})\}_{i=1}^{N}$ into a dataset of propositional symbols $\{(\Sigma_p^{(i)}, \Sigma_r^{(i)}, \mathbf{a}^{(i)}, \Sigma_p'^{(i)}, \Sigma_r'^{(i)})\}_{i=1}^{N}$ where
\begin{align*}
    \Sigma_p^{(i)} &= \{\sigma_p(\mathbf{x}_1^{(i)}), \dots, \sigma_p(\mathbf{x}_n^{(i)})\}\\
    \Sigma_r^{(i)} &= \{\sigma_r(\mathbf{x}_1, \mathbf{x}_1), \dots, \sigma_r(\mathbf{x}_i, \mathbf{x}_j), \dots, \sigma_r(\mathbf{x}_n, \mathbf{x}_n)\}
\end{align*}
and $\Sigma_p'^{(i)}$, $\Sigma_r'^{(i)}$ are defined similarly. For notational simplicity, we consider only a single relation type $\sigma_r$ while the same procedure can be applied to multiple relation types.

Our main goal is to  find a set of operators $\Phi=\{\phi_1, \dots, \phi_m\}$ parameterized by lifted actions $\alpha$ that are in the following form:
\begin{equation}
    \phi_i(\Sigma_p, \Sigma_r; \alpha_i) = (\Sigma_p', \Sigma_r')
\end{equation}
modeling the symbolic transition between states. We start by partitioning samples by their actions $\mathbf{a}^{(i)}$ and preconditions $\Sigma_p^{(i)}$, $\Sigma_r^{(i)}$: samples are grouped if their lifted actions and preconditions can be represented by the same substitution $\theta$. For example, consider the following samples:
\begin{align*}
    \Sigma_p^{(1)} &= \{\sigma_{p}(\mathbf{x}_1)=0, \sigma_{p}(\mathbf{x}_2)=0, \sigma_{p}(\mathbf{x}_3)=1 \} \\
    \mathbf{a}^{(1)} &= \text{pick-place}(\mathbf{x}_3, \mathbf{x}_1) \\
    \Sigma_p^{(2)} &= \{\sigma_{p}(\mathbf{x}_1)=0, \sigma_{p}(\mathbf{x}_2)=1, \sigma_{p}(\mathbf{x}_3)=0 \} \\ 
    \mathbf{a}^{(1)} &= \text{pick-place}(\mathbf{x}_2, \mathbf{x}_3)
\end{align*}
These samples can be grouped into the same category $C_1$ with substitutions $\theta_1=\{?a/\mathbf{x}_3, ?b/\mathbf{x}_1, ?c/\mathbf{x}_2\}$ and $\theta_2=\{?a/\mathbf{x}_2, ?b/\mathbf{x}_3, ?c/\mathbf{x}_1\}$. This procedure gives us a set of groups $\{C_1, \dots, C_k\}$ where each group is defined by lifted preconditions and actions. Next, we compute the lifted effects for each group:
\begin{align*}
    \mathcal{E}_p^{+} = \{\sigma \mid \sigma \in \Sigma_p'^{(i)}, \sigma \notin \Sigma_p^{(i)}\}\\
    \mathcal{E}_p^{-} = \{\sigma \mid \sigma \in \Sigma_p^{(i)}, \sigma \notin \Sigma_p'^{(i)}\}
\end{align*}
Relational effects $\mathcal{E}_r^{+}$ and $\mathcal{E}_r^{-}$ are computed similarly. If lifted effects are not the same for all samples in a group (i.e., a stochastic environment setting), we select the most frequent lifted effect for each group. This completes our operator definition:
\begin{align*}
    \phi_i(\Sigma_p, \Sigma_r; \alpha_i) &= (\Sigma_p', \Sigma_r') \\
    \Sigma_p' &= \Sigma_p \cup \mathcal{E}_p^{+} \setminus \mathcal{E}_p^{-} \\
    \Sigma_r' &= \Sigma_r \cup \mathcal{E}_r^{+} \setminus \mathcal{E}_r^{-}
\end{align*}

However, with this strategy, the number of groups increases with the number of objects. On the other hand, most of the time, only a subset of precondition symbols are relevant for a given action. For instance, if the action is to pick and place an object on top of another, then precondition symbols of other objects are irrelevant. Therefore, we only consider a subset of precondition symbols relevant to the action. Although determining which symbols are relevant is difficult to answer in a general sense, a practical and generally valid heuristic is to consider topological neighborhood or contact relations. In our experiments, we define this relevance as objects that are in the action arguments and objects that are in contact with these argument objects. A broader topological relevance alternative can also be to consider objects in the vicinity of action arguments.

\subsection{Translating Operators to PDDL}
\label{subsec:translating}
\begin{figure}[t]
    \centering
    \includegraphics[width=0.7\columnwidth]{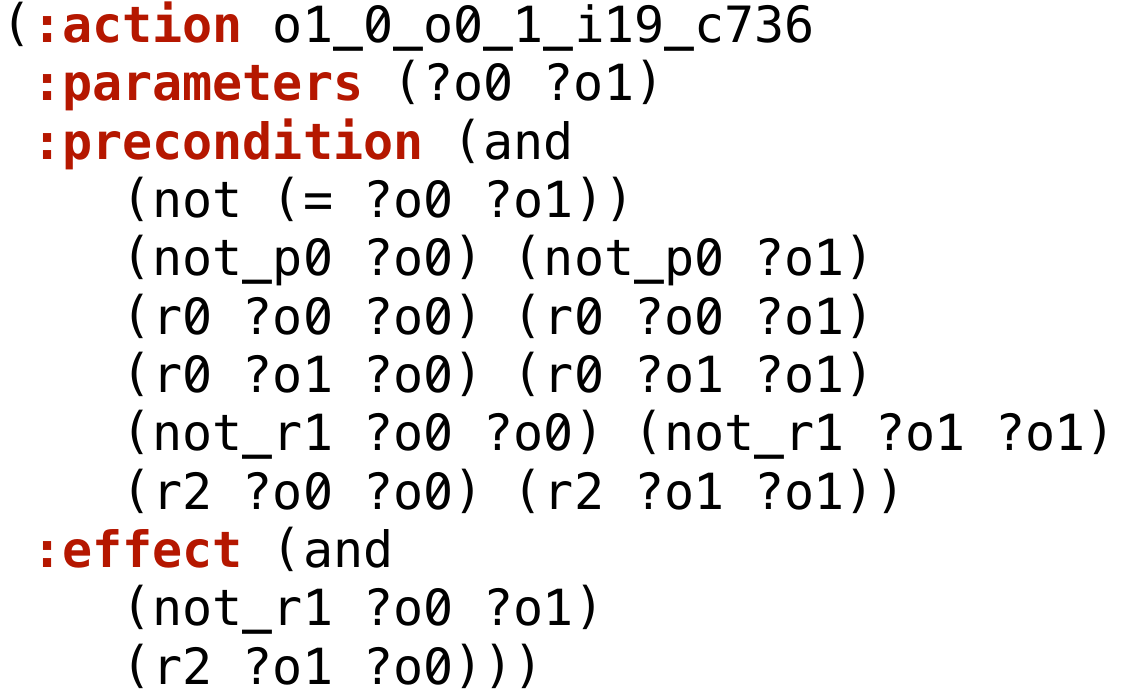}
    \caption{An example generated PDDL action schema. This action encodes pick-place($?o1$, `center', $?o0$, `right') where $?o0$ and $?o1$ are free variables. The action schema is the 20th most frequent action in the dataset, with 736 occurrences out of 160K samples.}
    \label{fig:pddl}
\end{figure}
Each operator $\phi_i$ is translated into a PDDL action schema where $\Sigma_p$ and $\Sigma_r$ are used as preconditions, $\mathcal{E}_p^{+}$ and $\mathcal{E}_r^{+}$ are used as effects, free variables that appear in the precondition and/or action are used as parameters, and the action name is defined by the action arguments. In the action schema, each $\sigma_{p_i}(?x)$ appear as \texttt{(p$i$ ?x)} or \texttt{(not\_p$i$ ?x)}, depending on the value of the predicate. We filter out action schemas that are used less than a threshold, which we set to 50 in our experiments. An example action schema is shown in Figure \ref{fig:pddl}. We observed that the most frequently used action schemas are empty actions, such as picking a short cube from the left, which results in no effect. This is due to the exploration process where we execute random actions in the environment that frequently result in no effect. Even though these definitions would not help in the planning process, we choose to keep them as they can be used in later exploration stages to avoid actions that do not have any effect and, thus, are not interesting for the agent.

\section{EXPERIMENTS}

\subsection{Experiment Setup}

We conducted our experiments in a tabletop object stacking environment (see top left in Figure \ref{fig:model}). There are two object types: 5x5x5 sized short block and 5x25x5cm sized long block. An environment instance contains two to four objects represented by pose and type. A UR10 robot arm has a single high-level action with four discrete parameters: picking an object from the left, right, or center of the object and releasing it on top of (or left of, right of) another object.

We followed the data collection procedure in \cite{ahmetoglu2023discovering}, where the robot executes random actions in the environment. The difference is that we only record objects that are either action arguments or in contact with them. These object features $\{\mathbf{x}_i\}_{i=1}^k$ are used as the state vector. The effect vector $\mathbf{e}_i$ for object $o_i$ is the difference between the next state $\mathbf{x}_i'$ and the current state $\mathbf{x}_i$. We subtract the lateral movement of the arm from the effect vector to remove the effect of the carry action. Otherwise, symbols need to encode global position information of objects, which is not necessary for planning since actions are already parameterized over objects. Note that state vectors can also be selected as raw pixels as in \cite{ahmetoglu2022deepsym,ahmetoglu2022learning}. We collect 200K samples and split them into 160K training, 20K validation, and 20K test samples.

We compare our method with \cite{ahmetoglu2022deepsym} and \cite{ahmetoglu2022learning} in terms of effect prediction accuracy, and with \cite{ahmetoglu2022learning} for planning performance. The original DeepSym formulation uses a fixed-sized vector to represent the environment's state, making it hard for us to compare the performance of the two methods. However, the comparison with \cite{ahmetoglu2022learning} already serves as a valuable measure to assess whether explicitly learning relational symbols helps the planning performance.

We train all methods for 4000 epochs with a batch size of 128 and a learning rate of 0.0001 using Adam optimizer \cite{kingma2014adam}. MLP blocks consist of two layers with 128 hidden units. In Relational DeepSym (ours), we set the number of relation types to three and the object symbol dimension to one (i.e., the encoder's output dimensionality). However, for other baselines, we set the object symbol dimension to four since these methods need more representational capacity to encode the state of the environment without relational symbols. Also, the number of attention heads for the attentive formulation \cite{ahmetoglu2022learning} is set to four. Following \cite{salimans2016weight}, for layers before GS activation, we normalize both the input and the weight vectors to have a norm of three. This prevents the vanishing gradients in the GS function. Lastly, we clip gradients by their norm to 10.

\subsection{Effect Prediction Results}
\begin{figure}[t]
    \centering
    \includegraphics[width=0.8\columnwidth]{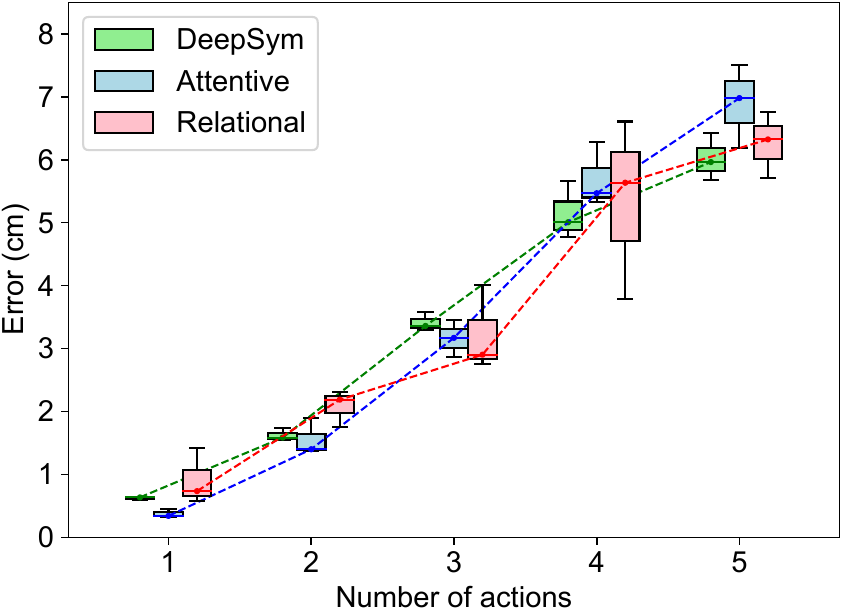}
    \caption{Effect prediction errors for different numbers of actions.}
    \label{fig:effres}
\end{figure}
\begin{table}[!t]
    \caption{Effect prediction results averaged over three runs. Units are in centimeters.}
    \begin{center}
    \begin{tabular}{ccc}
    \hline
    DeepSym & Attentive & Relational\\
    \hline
    4.79 $\pm$ 0.12 & 4.47 $\pm$ 0.10 & \textbf{3.21 $\pm$ 0.30}\\
    \hline
    \end{tabular}
    \label{tab:results}
    \end{center}
\end{table}
We report the test set effect prediction results in Table \ref{tab:results}. Relational DeepSym performs better than other methods by a small margin, which aligns with the results in \cite{ahmetoglu2023discovering}. In Figure \ref{fig:effres}, we compare these methods by their cumulative effect prediction error when predicting a sequence of actions by feeding the prediction back into the state in an autoregressive fashion. Even though we do not see a significant difference between different methods, this does not directly translate to planning performance. Attentive DeepSym uses transformer layers to pass information between object symbols, and there is no straightforward way of translating the relational knowledge embedded in the transformer weights into lifted operators. This is the key advantage of the Relational DeepSym as it directly encodes relations symbols together with object symbols.

\begin{figure*}[htbp]
    \centering
    \begin{subfigure}[b]{0.32\textwidth}
        \centering
        \includegraphics[width=\textwidth]{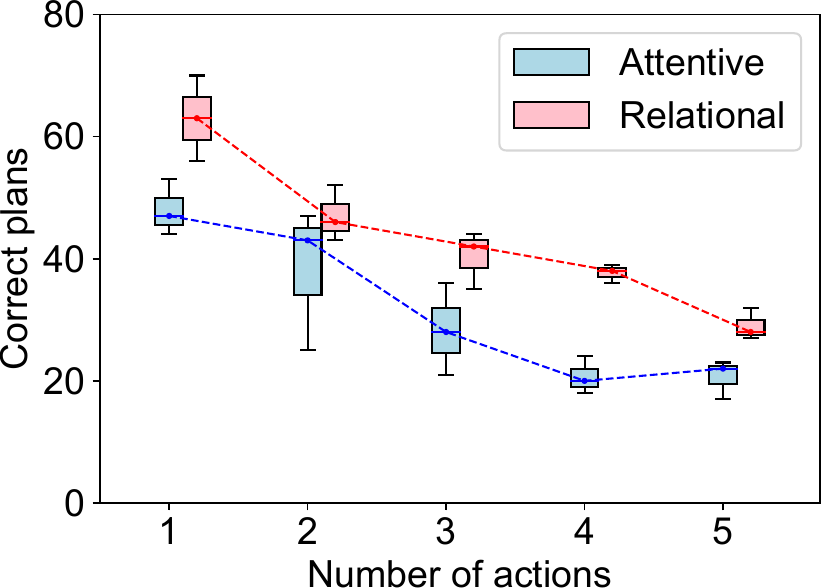}
        \caption{2 objects}
        \label{subfig:cum2obj}
    \end{subfigure}
    \hfill
    \begin{subfigure}[b]{0.32\textwidth}
        \centering
        \includegraphics[width=\textwidth]{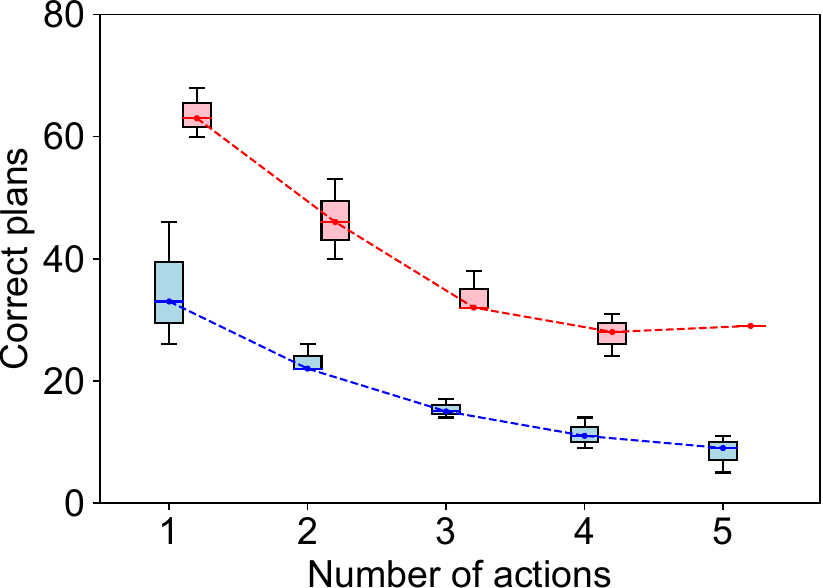}
        \caption{3 objects}
        \label{subfig:cum3obj}
    \end{subfigure}
    \hfill
    \begin{subfigure}[b]{0.32\textwidth}
        \centering
        \includegraphics[width=\textwidth]{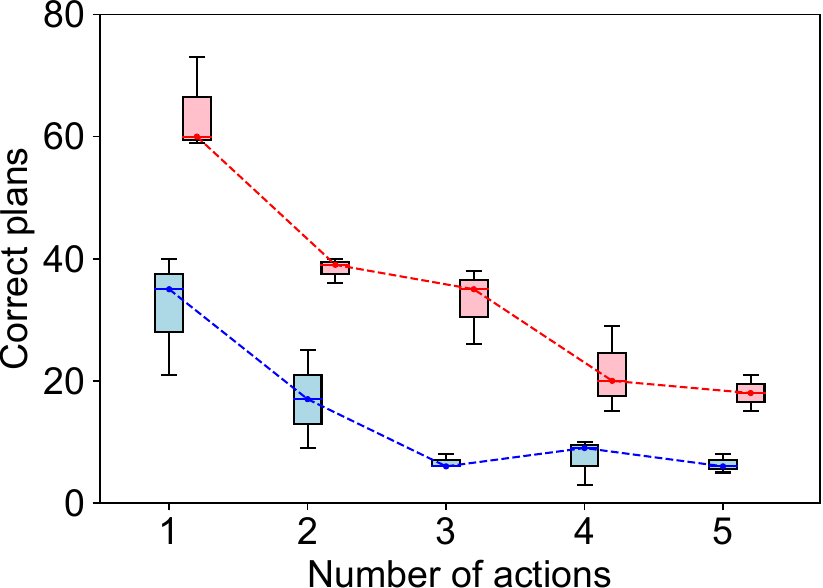}
        \caption{4 objects}
        \label{subfig:cum4obj}
    \end{subfigure}
    \caption{The planning performance for different numbers of objects over three runs with 100 random problem pairs.}
    \label{fig:errors}
\end{figure*}
\begin{figure*}[t]
    \centering
    \includegraphics[width=\textwidth]{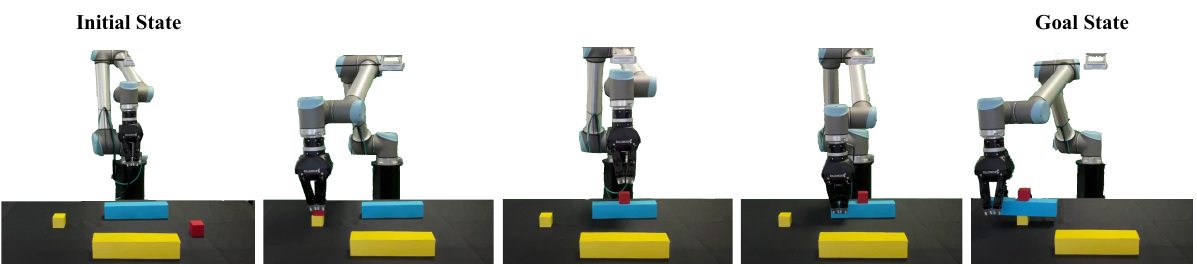}
    \caption{Given an initial environment configuration in the first column, our model can generate an action sequence reaching the goal state.}
    \label{fig:real}
\end{figure*}

\begin{figure}[htbp]
    \centering
    \begin{subfigure}[b]{0.48\columnwidth}
        \centering
        \includegraphics[width=\textwidth]{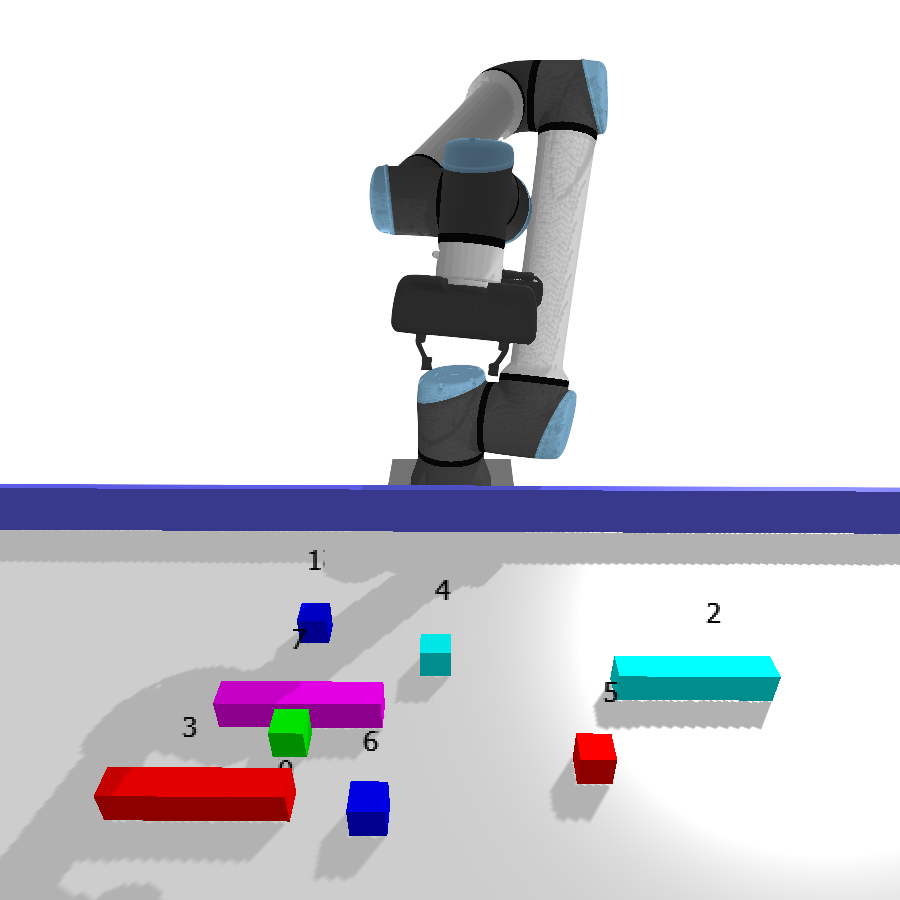}
    \end{subfigure}
    \hfill
    \begin{subfigure}[b]{0.48\columnwidth}
        \centering
        \includegraphics[width=\textwidth]{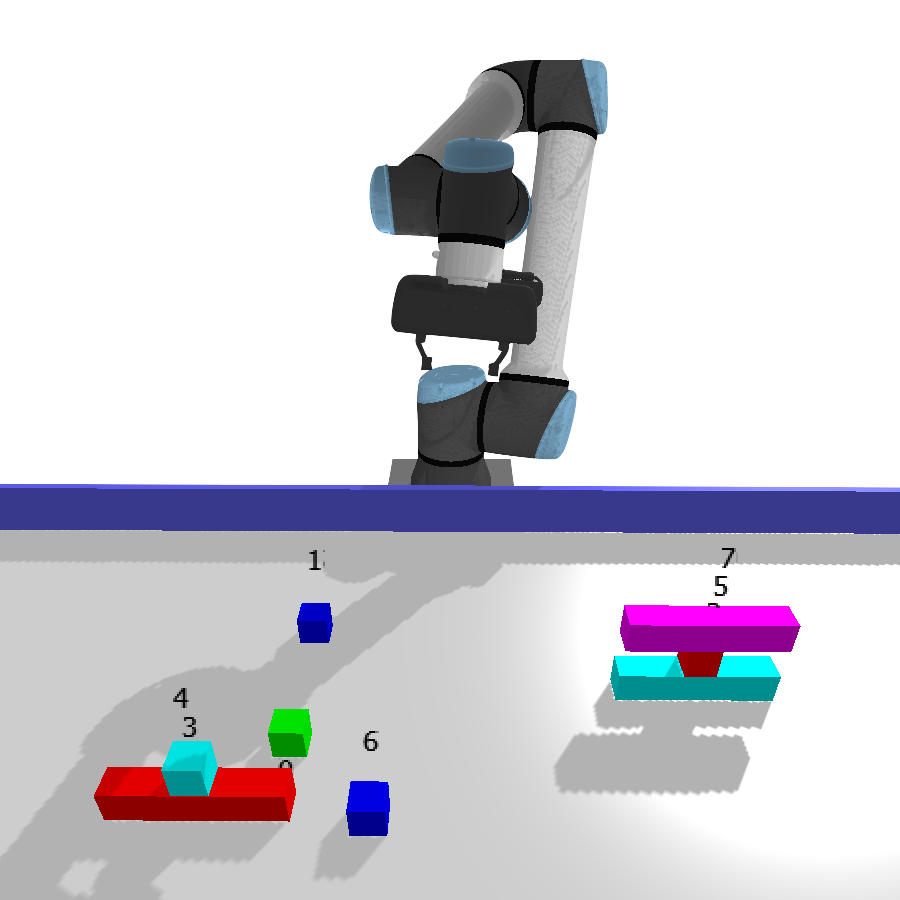}
    \end{subfigure}
    \caption{Given an initial state (left) and a goal state (right), Relational DeepSym can find a plan to achieve the goal even though it is only trained with two to four objects.}
    \label{fig:plan}
\end{figure}

\subsection{Planning Performance}
In this section, we compare the planning performance of Relational DeepSym with the attentive formulation \cite{ahmetoglu2022learning}. We generate a random set of problem pairs $\{(\mathbf{X}_0^{(i)}, \mathbf{X}_g^{(i)})\}_{i=1}^{N}$ by executing random actions on the environment. For each problem pair, we convert $\mathbf{X}_0$ and $\mathbf{X}_g$ into their symbolic counterparts $\Sigma_{p_0}$, $\Sigma_{r_0}$ and $\Sigma_{p_g}$, $\Sigma_{r_g}$, and produce PDDL problem statements. We filter out relations for object pairs that are not in contact in the goal state; otherwise, the planner might fail to find a plan due to spurious relations. We use the Fast Downward planning system \cite{helmert2006fast} and set a timeout limit to 10 seconds. We automatically check whether a plan is correct by computing cartesian distances of objects from the goal state and accept it as correct if the difference is less than 5cm for all objects.

Figure \ref{fig:errors} shows the planning performance for different numbers of objects and actions. Planning with the domain defined over unary and relational symbols generated by Relational DeepSym performs significantly better than the implicit attentive version in which all relational information remains hidden in the network's weights. We also test our method in a real-world experiment with the same environment definition (Figure \ref{fig:real}). We detect the locations of objects with an Intel Realsense depth camera by clustering pixels and transform them into the robot frame. We manually define a goal configuration with its corresponding contact graph. The rest of the algorithm works the same as in simulation experiments.

Another advantage of using relational symbols in addition to unary symbols is that we can represent an environment configuration that has many more objects than the ones in the training set since the number of objects does not affect the encoded representation. This is not the case for the attentive formulation \cite{ahmetoglu2022learning} as the model is biased towards the number of tokens in the training set, similar to other transformer-based architectures. In Figure \ref{fig:plan}, we initialize the scene with eight objects and create a goal configuration that was not in the train set. The planner successfully finds an action sequence that reaches the goal state.

\section{DISCUSSION}
In our experiments, we define \emph{relevance} as objects that are in the action arguments and objects that are in contact with these argument objects. An alternative relevance might be to consider objects in the vicinity of action arguments. Although this creates a limitation, most of our daily activities can be represented with this relevance. An example exception is controlling a race car with a joystick; learning the relevance between such objects should be treated as a separate problem.

% The operator learning procedure defined in Section \ref{subsec:ops} is similar to \cite{chitnis2022learning} in which operators are learned by partitioning samples by their lifted actions and effects instead of preconditions. In our early experiments, we observed that when there are some erroneous symbolic transitions in the dataset (due to an erroneous prediction by the network or stochasticity of the environment) that are not consistent with the rest of the dataset, partitioning by effect may result in operators with erroneous preconditions.

\section{CONCLUSION}
In this study, we proposed and implemented a framework where object and relational predicates are discovered from the continuous sensory experience of the robot, symbolic rules that encode the transition dynamics that involve pick and place actions on arbitrary numbered compound objects are extracted, these rules are automatically transferred to PDDL and symbolic plans are generated and executed to achieve various goals. We showed that the planning performance of our framework significantly outperforms the baselines. The key factor for the performance increase is representing the environment with relational symbols in addition to object symbols. In future work, we plan to use the learned operators to search for novel states and effects, which would drastically increase the sample efficiency and thus, the learning progress.

\addtolength{\textheight}{-6cm}   % This command serves to balance the column lengths

\section*{ACKNOWLEDGMENT}
This research was supported by TUBITAK (The Scientific and Technological Research Council of Turkey) ARDEB 1001 program (project number: 120E274). Additional support was given by the Grant-in-Aid for Scientific Research (project no 22H03670), the project JPNP16007 commissioned by the New Energy and Industrial Technology Development Organization (NEDO), JST, CREST (JPMJCR17A4), and by INVERSE project (no. 101136067) funded by the European Union.

\bibliographystyle{IEEEtran}
\bibliography{IEEEabrv,ref}

\end{document}